\documentclass[letterpaper]{article} 
\usepackage{aaai25}  
\usepackage{times}  
\usepackage{helvet}  
\usepackage{courier}  
\usepackage[hyphens]{url}  
\usepackage{graphicx} 
\usepackage{natbib}  
\usepackage{caption} 
\usepackage{algorithm}
\usepackage{algorithmic}

\usepackage{newfloat}
\usepackage{listings}
\DeclareCaptionStyle{ruled}{labelfont=normalfont,labelsep=colon,strut=off} 
\floatstyle{ruled}
\newfloat{listing}{tb}{lst}{}
\floatname{listing}{Listing}
\usepackage{multirow}
\usepackage{pifont}
\usepackage{subcaption} 
\usepackage{booktabs}
\usepackage{amsmath}

\title{Video Repurposing from User Generated Content: \\ A Large-scale Dataset and Benchmark}
\author{
    Yongliang Wu\textsuperscript{\rm 1,2}\equalcontrib, Wenbo Zhu\textsuperscript{\rm 2}\equalcontrib\thanks{Project leader}, Jiawang Cao\textsuperscript{\rm 2}\equalcontrib, Yi Lu\textsuperscript{\rm 2,3}, Bozheng Li\textsuperscript{\rm 2,4}, \\ Weiheng Chi\textsuperscript{\rm 2,5}, Zihan Qiu\textsuperscript{\rm 2}, Lirian Su\textsuperscript{\rm 2}, Haolin Zheng\textsuperscript{\rm 2}, Jay Wu\textsuperscript{\rm 2}, Xu Yang\textsuperscript{\rm 1}\thanks{Corresponding author}\\
}
\affiliations{
    \textsuperscript{\rm 1}Southeast University, \textsuperscript{\rm 2}Opus AI Research, \textsuperscript{\rm 3}University of Toronto\\
     \textsuperscript{\rm 4}Brown University, \textsuperscript{\rm 5}National University of Singapore\\

    \{yongliangwu, xuyang\_palm\}@seu.edu.cn
}

\usepackage{bibentry}

\begin{document}

\maketitle

\begin{abstract}
The demand for producing short-form videos for sharing on social media platforms has experienced significant growth in recent times. Despite notable advancements in the fields of video summarization and highlight detection, which can create partially usable short films from raw videos, these approaches are often domain-specific and require an in-depth understanding of real-world video content. To tackle this predicament, we propose Repurpose-10K, an extensive dataset comprising over 10,000 videos with more than 120,000 annotated clips aimed at resolving the video long-to-short task. Recognizing the inherent constraints posed by untrained human annotators, which can result in inaccurate annotations for repurposed videos, we propose a two-stage solution to obtain annotations from real-world user-generated content. Furthermore, we offer a baseline model to address this challenging task by integrating audio, visual, and caption aspects through a cross-modal fusion and alignment framework. We aspire for our work to ignite groundbreaking research in the lesser-explored realms of video repurposing.
\begin{links}
    \link{Code}{https://github.com/yongliang-wu/Repurpose}
\end{links}
\end{abstract}

\section{Introduction}
\label{sec:intro}
Driven by the rapid growth of social media platforms such as Instagram, TikTok, and YouTube Shorts, short-form videos become the primary medium for sharing daily life and conveying information. Even creators who traditionally focus on long-form content, such as live streams, interviews, and vlogs, now shift towards producing captivating short-form videos for these platforms~\cite{wang2023podreels}. This shift underscores the importance of identifying the most engaging clips within longer videos while maintaining logical coherence.

This brings us to the critical task of video repurposing~\cite{singh2004guest}. We introduce the task of long-to-short video repurposing, a sophisticated process that transforms user-generated videos, typically longer than 30 minutes, into a series of engaging clips around 60 seconds each, referred to as repurposed clips. This process involves extracting and highlighting key segments to ensure the overall narrative remains complete and engaging. It includes setting a length limit for the final video and re-editing or reorganizing the retrieved content to form a coherent and engaging clip. The goal is to produce a compelling narrative suitable for direct publication on social media platforms.

\begin{figure*}[t]
    \centering
    \includegraphics[width=\linewidth]{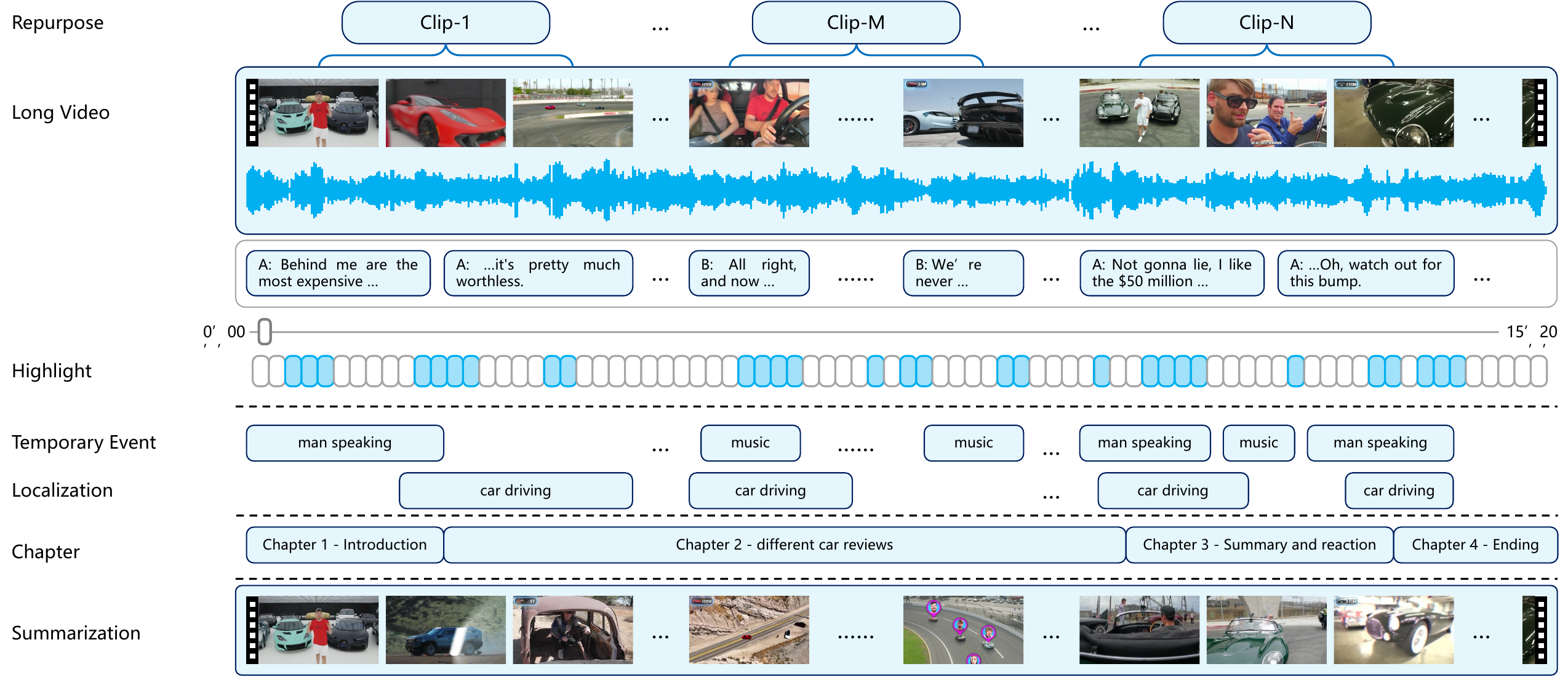}
    \caption{The distinction between Video Repurposing and other similar tasks. From top to bottom: Video Repurposing, Highlights Detection, Temporary Event Localization, Video Chapter, and Video Summarization.}
    \label{fig:teaser}
\end{figure*}

The challenge of automatically repurposing videos remains an unresolved issue. Existing methods, such as shot boundary detection (SBD)~\cite{zhu2023autoshot}, temporal event localization (TEL)~\cite{geng2023dense,wu2024number}, video chapter~\cite{yang2024vidchapters}, and temporal action detection (TAD)~\cite{yang2023basictad}, serve as auxiliary tools for video repurposing but do not fully address the core problem. Two other related tasks are video highlight detection~\cite{badamdorj2022contrastive} and video summarization~\cite{zhang2016video,hu2023dual}. The former involves autonomously identifying the most captivating moments within a raw video, while the latter aims to create a concise synopsis that succinctly summarizes the video content by selecting its most informative and pivotal segments. However, it is important to note that previously proposed datasets and methodologies are often domain-specific and do not necessarily meet the unique requirements of video repurposing, which demands a comprehensive understanding of the content~\cite{sun2014ranking, potapov2014category, panda2017weakly, song2015tvsum, pei2022global,wang2022text,wu2023empirical}. The difference between these tasks is shown in Figure~\ref{fig:teaser}.

Based on this, we compile a large-scale dataset named Repurpose-10K, which includes more than 10,000 video samples and more than 120,000 annotated clips, making it a formidable benchmark for video repurposing. Previous research~\cite{demartini2021managing} highlights the inherent limitations of traditional reliance on human annotators, such as the introduction of bias and potential inaccuracies from untrained individuals. Given the challenges associated with precise clip annotation in videos, we develop an approach that involves acquiring annotations from authentic User Generated Content (UGC) in the real world. The annotation methodology consists of three distinct stages. Initially, repurposing tools based on large language models are utilized to transform long-form videos into shorter clips, establishing preliminary edits with coarse-cut boundaries. Users then express their preferences for each clip by marking them as ``like'' or ``dislike''. Clips marked as ``like'' are selected from the video. Finally, content creators are invited to meticulously refine the start and end timestamps for each clip through manual annotation, resulting in the creation of the final annotated dataset.

To establish robust baseline models to address this challenge, we introduce an end-to-end transformer encoder-decoder architecture that conceptualizes video repurposing as a joint classification and regression problem~\cite{geng2023dense,li2024laso,li2023transformer,shi2024LLMFormer,shi2023unified}. To fully leverage the information from all the modalities within the video, we introduce the Caption Enhancement Module. This module utilizes captions generated by an Automatic Speech Recognition (ASR) model to aid in the alignment of visual and audio modalities. Given that the model may rely on information from a single modality to make judgments, for instance, the audio branch might select segments with engaging music, while the visual branch might choose visually striking segments, relying solely on fused features can lead to insufficient integration of modal features. To address this, we propose the Multi-Modal Align Guider. This module is designed to guide each modality towards selecting the appropriate segments, ensuring consistency and alignment across the different modalities. Through comprehensive experimental analysis and comparison with other state-of-the-art video highlight detection models, we demonstrate the superior performance and practical applicability of our proposed model for this task.

Our contributions are as follows: (1) We introduce Repurpose-10K, a large-scale dataset specifically curated for the video repurposing task. Unlike traditional crowd-sourcing methods for annotation, Repurpose-10K compiles annotations from user-generated content (UGC). To the best of our knowledge, this is the first large-scale dataset developed specifically for the challenging video repurposing task. (2) We propose a baseline model tailored for the video repurposing task. This model effectively integrates multi-modal information from videos and includes the design of a Multi-Modal Align Guider module, which aims to maximize the agreement and ensure consistency across the different modalities. (3) Through a comprehensive experimental analysis, we validate the effectiveness and practical feasibility of our model. The results demonstrate superior performance and practical applicability in the task of video highlight detection.

\section{Related Work}
\label{sec:related_work}
\subsection{UGC Video Long to Short}
In recent years, video podcasts surge in popularity~\cite{wang2023podreels}. Platforms like Instagram Reels, TikTok, and YouTube Shorts offer new opportunities to enhance podcast visibility. Successful podcasters often use teasers, brief highlights from their episodes, to attract new listeners. The strategy of repurposing, which involves editing and reshaping existing videos for different objectives or audiences, becomes a prominent method for converting long-form content into short, digestible formats. This technique helps viewers to assimilate the information presented in the videos more efficiently~\cite{truong2021automatic}. The key to repurposing content lies in identifying valuable segments within a video that meet the criteria for short-form video platforms. While previous video annotations, such as highlights~\cite{sun2014ranking, lei2021detecting}, focus on the visual appeal of images, they often do not require high integrity of user-generated content and are thus unsuitable for direct repurposing. Addressing video editing tasks that demand deeper semantic understanding reveals a significant gap in large-scale, labeled datasets and end-to-end methods.

\subsection{Related Tasks}
\paragraph{\textbf{Temporal Event Localization.}}
Temporal Event Localization (TEL)~\cite{tian2018audio, tian2020unified, lee2021acav100m, geng2023dense,zhang2024causal1,li2024frame,zhang2024causal2} is a video understanding task that aims to identify and pinpoint the start and end times of events or actions of interest within untrimmed videos. This task proves crucial for areas such as multimedia content analysis, video surveillance, and sports event analysis. The supervised learning paradigms of TEL are categorized into two primary methodologies: two-stage approaches~\cite{zhao2017temporal,li2023adversarial,li2024two} and single-stage frameworks~\cite{buch2019end, zhang2022actionformer, yan2023unloc,li2024self,liu2024omniclip}. Recently, UnLoc~\cite{yan2023unloc} proposes a unified framework for video localization tasks, consisting of a two-tower CLIP model, with output features fed into a video-text fusion module and feature pyramid. Unlike TEL, which typically focuses on identifying explicit timestamps, video repurposing revolves around identifying complete and engaging segments within lengthy videos, without explicitly marking a specific starting point, as a temporal localization task~\cite{liu2024envisioning}.

\paragraph{\textbf{Video Highlight Detection.}}
Video highlight detection~\cite{sun2014ranking} aims to automatically identify the most compelling and quintessential segments of a video. While earlier methods consider only visual features~\cite{sun2014ranking}, recent works start to incorporate both audio and visual components to retrieve highlights from the video~\cite{badamdorj2021joint}. Query-Dependent DETR~\cite{moon2023query} employs a cross-attention transformer encoder to generate more potent multi-modal video representations, deliberately infusing text queries into the feature representation. However, highlight detection particularly emphasizes moments that are exceptionally striking or exhilarating, rather than necessarily providing a coherent summary of the video content. For example, a moment is considered a highlight if at least 70\% of its frames match the edited video in the training of YouTube Highlights~\cite{sun2014ranking}. This approach to combining highlights poses challenges in crafting a comprehensive narrative thread, making it difficult for direct publication on short-form social media platforms due to the lack of a cohesive storyline. Additionally, repurposing video highlights often requires post-processing to form coherent segments, introducing further effort.

\paragraph{\textbf{Video Summarization.}}
Video summarization~\cite{song2015tvsum} aims to extract key information from lengthy video content and generate a concise summary version. This task preserves the most informative and representative parts of the video, allowing users to quickly grasp the main content without watching the entirety of the original video. The key parts within a video could either be a series of key frames or a collection of one or more key segments. To better capture temporal correlations in video content, a variety of sophisticated deep learning models~\cite{zhang2016video, li2022video} develop to map dependencies across time, using either localized or holistic approaches. Video summarization prioritizes comprehensiveness, often overlooking the engaging aspect of video content. In contrast, video repurposing does not mandate conveying the entire essence of the video.

\begin{table}[t]
    \centering
    \setlength\tabcolsep{1mm} 
    \fontsize{9}{11}\selectfont 
    \begin{tabular}{@{}lcccccc@{}}
        \toprule
        Dataset & Domain & Sizes & Avg. Length & Modality & TB \cr
        \midrule
        AVE & TEL & 4,143 & 10s & AV & \ding{51} \\
        LLP & TEL & 11,849 & 10s & AV & \ding{51} \\
        ACAV100M & TEL & 100M & 10s & AV & \ding{55} \\
        UnAV-100 & TEL & 10,790 & 42.1s & AV & \ding{51} \\
        \hline  
        OVP & VS & 50 & 1.5 mins & AV & \ding{55} \\
        SumMe & VS & 25 & 2.4 mins & AV & \ding{55} \\
        TVSum & VS & 50 & 4.2 mins & AV & \ding{55} \\
        \hline
        YT-Highlights & VH & 433 & 143s & AV & \ding{51} \\
        QVHighlights & VH & 10,148 & 150s & AV & \ding{51} \\
        \hline
        MultiSeg & TSLLV & 1,000 & 78 mins & AVC & \ding{51} \\
        \hline
        Repurpose-10K & VR & 11,210 & 32 mins & AVC & \ding{51} \\
        \bottomrule
    \end{tabular}
    \caption{Comparison with related multi-modal video datasets. A: audio; V: visual; C: caption; TB: temporal boundaries; TEL: Temporal Event Localization; VH: Video Highlights; VS: Video Summarization; TSLLV: Temporal Segmentation of the Long Livestream Videos; VR: Video Repurposing.}
    \label{tab:datasets}
\end{table}

\section{Repurpose-10K Dataset}
\subsection{Overview}
We propose the Repurpose-10K dataset. By leveraging a repurposing SaaS platform, we collect annotations from real users. For each long video, the annotations include timestamps marking the start and end of multiple clips, with each clip encapsulating a self-contained sub-topic, such as a compilation of actions or a dialogue exchange.

\begin{figure*}[t]
    \centering
    \begin{minipage}[t]{0.33\linewidth} 
    \centering
    \includegraphics[scale=0.28]{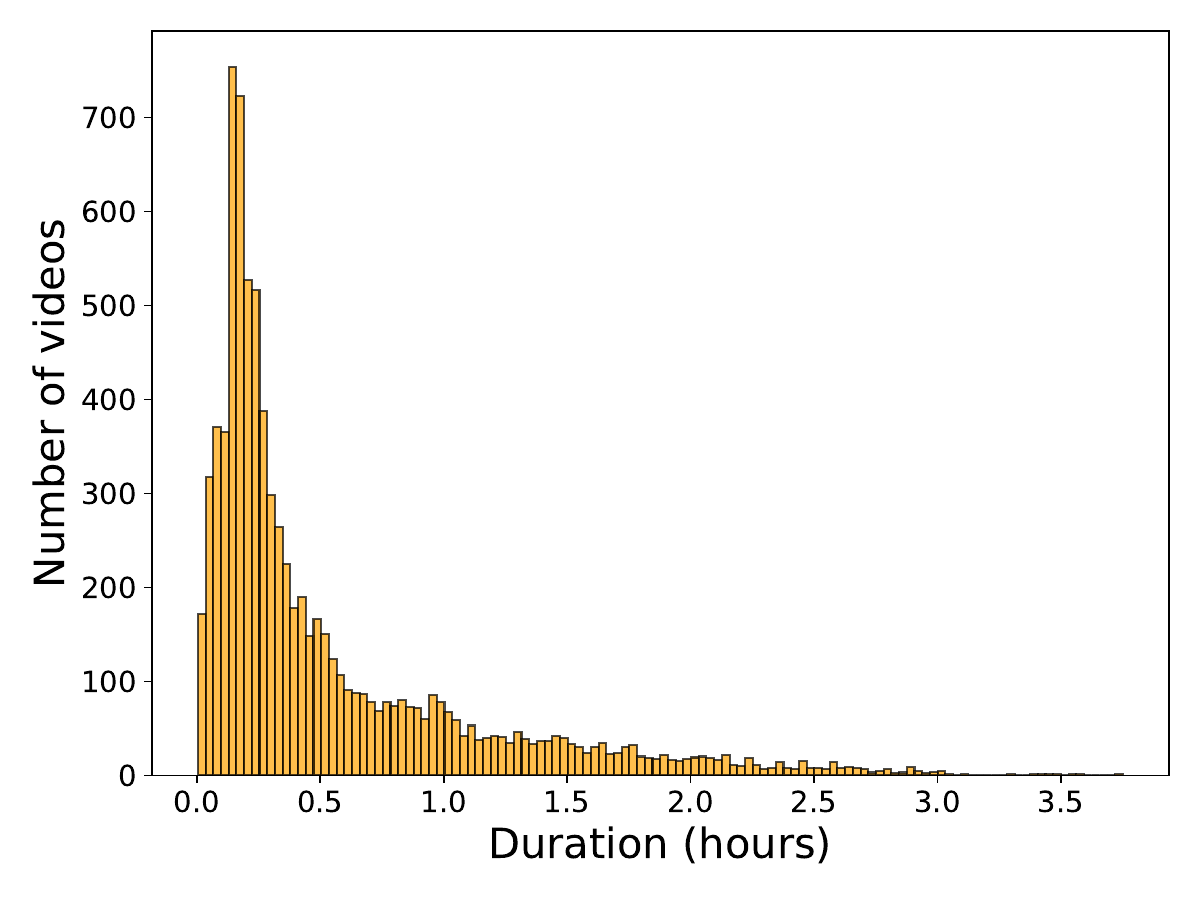}
    \end{minipage}
    \begin{minipage}[t]{0.33\linewidth} 
    \centering
    \includegraphics[scale=0.28]{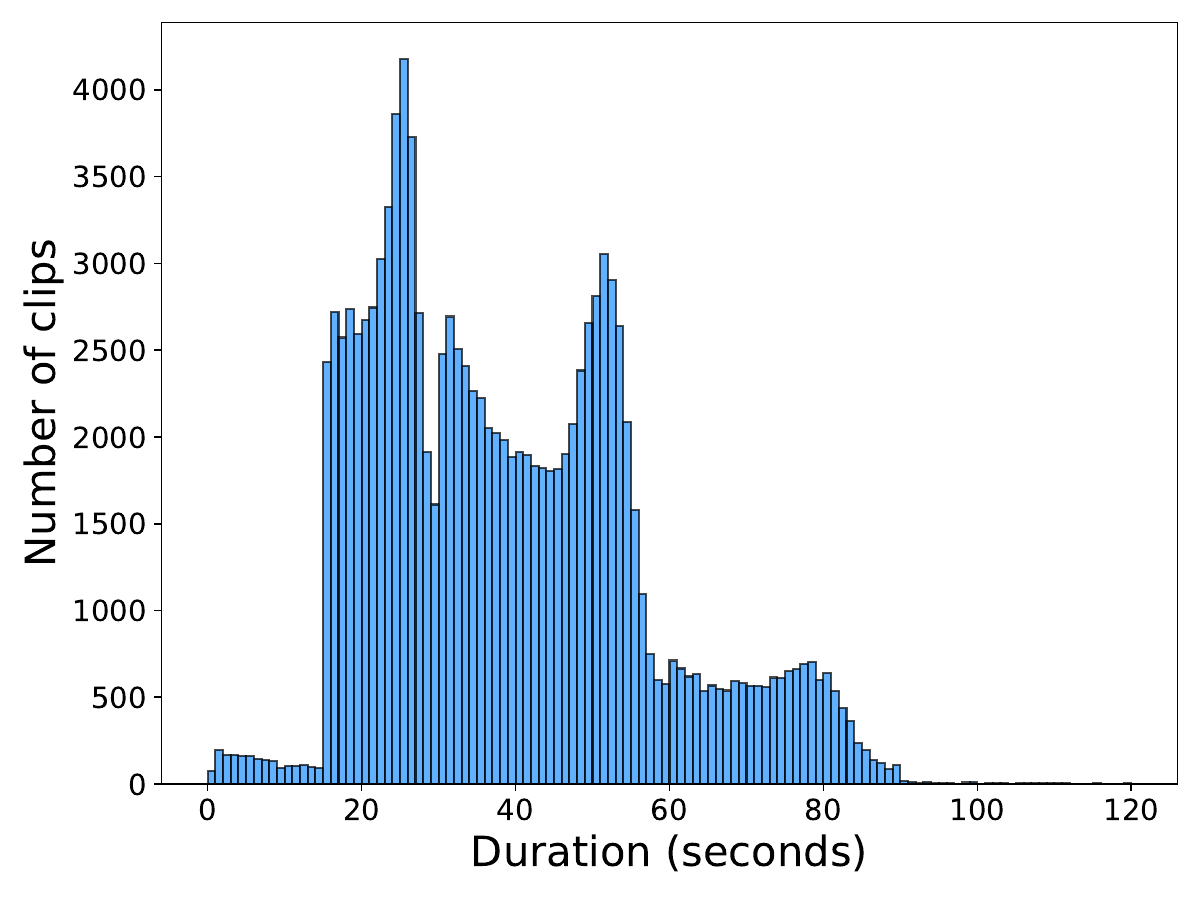}
    \end{minipage}
    \begin{minipage}[t]{0.33\linewidth} 
    \centering
    \includegraphics[scale=0.28]{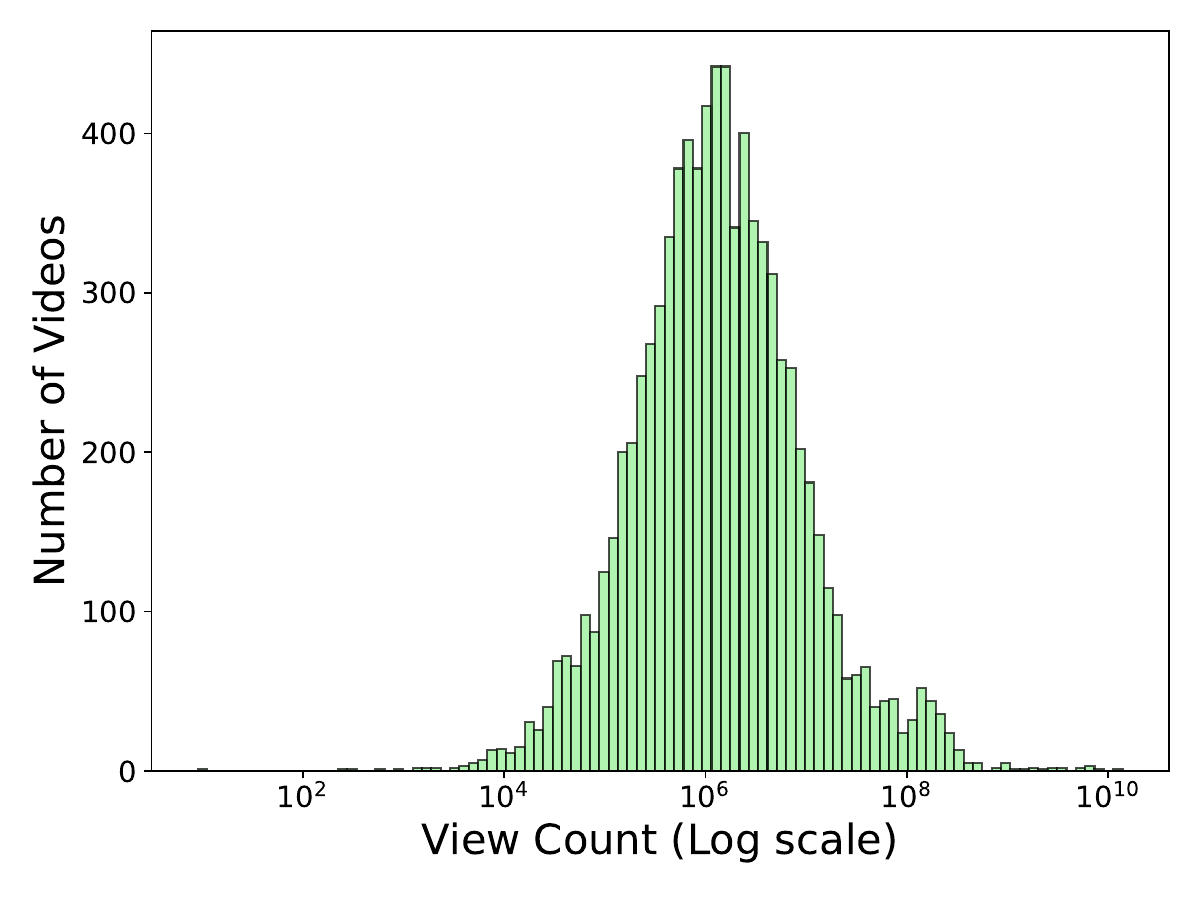}
    \end{minipage}
    \caption{Histogram of Repurpose-10K videos duration. From left to right: collection videos duration, repurpose clips duration (y-axis: number of videos), log-scale distribution of collection videos view counts.}
    \label{fig:combined}
\end{figure*}

The comparison of our dataset with related datasets is presented in Table \ref{tab:datasets}. We argue that the quantity of videos plays a critical role in User Generated Content (UGC) video understanding tasks. When comparing video summarization datasets, their quantities are relatively small, which may not sufficiently represent the diversity and complexity needed for repurposing challenges. Furthermore, addressing the challenges posed by real-world repurposing tasks, which involve the processing of untrimmed videos, we provide an extensive collection of long-format videos, averaging 32 minutes each. Additionally, given the prevalence of captions, we include them as part of the dataset to facilitate a comprehensive understanding of the multi-modal content.

\subsection{Data Collection}
To gain a deeper insight into user preferences regarding UGC, we conduct an extensive collection of real-world user interaction data. Initially, leveraging the topic segmentation capabilities of the OPUS platform\footnote{https://www.opus.pro/}, a versatile video repurposing tool, users divide YouTube videos into multiple short sections related to various topics, creating a preliminary edit with coarse-cut boundaries. Then, these users personally select their preferred clips from the entire collection of these sections by clicking the ``like" or ``post" button. We crawl 8,398 YouTube videos posted by 114,883 content creators. In these videos, there are at least one or more repurposed clips that users mark as ``like" or ``post". To ensure the accuracy of the timestamps, we selectively choose clips that have undergone user re-editing for the start and end timestamp refinement, resulting in 120,925 finely annotated video clips. Strictly limited to clips that are manually annotated by users and subsequently endorsed as their preferred choices, our dataset comprises a curated selection of user-preferred content.

\subsection{Data Analysis}
The Repurpose-10K dataset includes an extensive collection of 8,398 videos. The total duration of these videos amounts to 4,539.94 hours, with an average length of 0.54 hours per video. To manage the annotation workload, videos exceeding 30 minutes are divided into multiple clips, resulting in a total of 11,210 videos. We also compile view counts using yt-dlp\footnote{https://github.com/yt-dlp/yt-dlp}, with an average of 18.9 million views per video, and a 90\% confidence interval ranging from 62,900 to 38.6 million views. The distribution, illustrated in Figure~\ref{fig:combined}, suggests that these long videos have inherent dissemination value, making them suitable for repurposing tasks.

\subsection{Human Evaluation}
To assess the annotation accuracy in video repurposing within UGC content, we conduct a comparative user study on various repurposing outcomes. We randomly select a sample of 110 videos, yielding a total of 462 labeled results with an average duration of 58.66 seconds. Additionally, we extract 462 clips with random timestamps from the original videos, each ranging from 30 to 90 seconds. Two professional short video creators are also employed to identify the clips within the videos that hold the highest potential for repurposing, producing 3 short clips for each video. These clips are then randomized, and 10 participants are invited to partake in a survey to evaluate the clips. Participants cast votes and rate the clips from different perspectives (engagement level and completeness level, each ranging from 1 to 5). The conclusive scores are presented in Table~\ref{tab:human_eva}. The results reveal a close alignment between user-generated labels and those provided by professional editors. Notably, the user-generated labels outperform randomly assigned clips by a significant margin.

\begin{table}[t]
    \centering
    \setlength\tabcolsep{0.5mm} 
    \fontsize{9}{11}\selectfont
    \begin{tabular}{ccccccccc}  
        \toprule
        \multirow{2}{*}{Group} & \multicolumn{3}{c}{Duration(s)} & \multicolumn{3}{c}{~Avg.Score} \\
        \cline{2-7}
        & Min & Max & Avg &  E & C & Total \\
        \hline
        \centering Random  & ~33.39 & ~88.90 & ~59.55 &  ~1.623 &  ~1.508 &  ~3.131 \\
        \centering Professional Editor   & ~31.93 & ~89.25 &  ~53.27 &  ~3.938  &  ~3.738 &  ~7.676 \\
         \centering Repurpose-10K  & ~20.42 & ~154.41 &  ~58.66 &  ~3.868  &   ~3.418 &   ~7.286 \\
	\bottomrule
    \end{tabular}
        \caption{Human evaluation results from different annotation. ( E: Engaging level, C: Completeness level ) }
            \label{tab:human_eva}
\end{table}

\section{Method}
\label{sec:method}
\subsection{Preliminary}
We approach video repurposing by jointly learning classification and regression tasks, taking video elements (segments, audio, captions) as input and output the time range for selected clips. The classification task determines if a segment should be selected, while the regression task pinpoints the exact temporal offsets for the clip's start and end. For an input video sequence with visual and audio tracks, we segment it into visual and audio pairs, represented as \(\{V_{t}, A_{t}\}_{t=1}^{T}\), where \(T\) is the number of segments. In cases involving human speech, we use whisperX~\cite{bain2022whisperx} to convert audio segments into time-stamped captions, \(C_t\), enhancing our segment pairs to \(\{V_{t}, A_{t}, C_t\}_{t=1}^{T}\). For each video, we select a set of clips, compiling temporal intervals denoted by \(R = \{(t_{s,n}, t_{e,n})\}_{n=1}^{N}\), where \(N\) is the number of curated excerpts.

In the classification task, segments within selected clips are labeled 1, and those outside are labeled 0, creating a binary classification framework. In the regression task, we simplify by calculating the temporal offset from the segment's timestamps to the start and end of the clip. This is expressed as \(Y = \{d_{s,t}, d_{e,t}, c_{t}\}_{t=1}^{T}\), where \(d_{s,t}\) and \(d_{e,t}\) are the temporal distances to the start and end of the clip, and \(c_{t}\) is the binary classification label (0 or 1). \(T\) represents the total number of labeled segments in the video.

\begin{figure*}[t]
    \centering
    \includegraphics[width=0.9\linewidth]{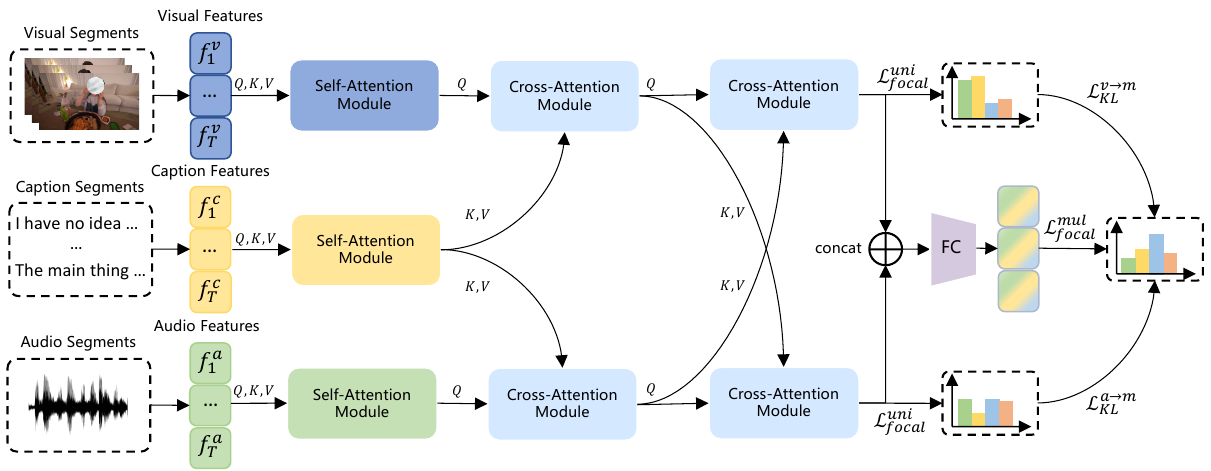}
    \caption{The overall architecture of our proposed baseline model consists of two main components: the Caption Enhancement Encoder and the Multi-Modal Align Guider. Q: Query. K: Key. V: Value. FC: Fully Connected Layer.}
    \label{fig:method}
\end{figure*}

\subsection{Framework}
\paragraph{\textbf{Feature Extractor.}} As shown in Figure~\ref{fig:method}, we begin by extracting three modalities of data using dedicated pre-trained models. This results in \(F_V = \{f_t^v\}_{t=1}^T\) for the visual modality, \(F_A = \{f_t^a\}_{t=1}^T\) for the audio modality, and \(F_C = \{f_t^c\}_{t=1}^T\) for the caption modality. We then use three distinct MLP layers to map these features to a unified dimension \(d\).

\paragraph{\textbf{Caption Enhancement Encoder.}} Initially, we employ three separate self-attention modules, each with \(N_s\) layers, to capture the temporal information within each modality. We treat caption features as an auxiliary component, noting the positive correlation between audio and caption features. Captions often explicitly describe spoken language and visually referenced objects, accompanied by corresponding audio cues. For example, if a car appears in the scene, the caption might describe it while the audio captures the engine's sound. We leverage caption features to align and discern visual and audio attributes. To achieve this, we introduce two cross-attention modules with \(N_c\) layers to facilitate information exchange and synergy among the modalities. Finally, we use \(N_f\)-layer cross-attention modules to promote dynamic interaction between visual and audio modalities. The fused features are then concatenated, mapped to lower dimensions, and fed into the decoder module.

\paragraph{\textbf{Multi-Modal Alignment Guider.}} The use of multi-modal data holds great potential for enhancing learning by integrating features from various domains, thereby improving model performance. However, the fusion process can sometimes be ineffective in bridging the gaps between different modalities, leading to suboptimal results. Direct fusion for predictions can introduce instability during the model's learning process. To address this issue, we introduce the Multi-Modal Alignment Guider. This module is designed to guide each modality in selecting appropriate segments, ensuring consistency and alignment across the different modalities. For each branch, we implement a classification head composed of a three-layer Multilayer Perceptron (MLP) with a sigmoid activation function to predict outcomes for the current video sequence. We assume that the uni-modal branches select segments based on the specific information available within their respective modalities. For instance, the audio branch may choose segments with compelling music, while the visual branch selects visually striking segments. In contrast, the multi-modal branch makes predictions based on the fused features of both modalities. During this process, we use focal loss~\cite{lin2017focal}, denoted as \(\mathcal{L}_{focal}\), to predict each segment individually. The classification loss can be formulated as follows:

\begin{equation}
\mathcal{L}_{\text {focal}}^{uni}=- (\sum_{i=1}^T {c}_i \log {y}_i^{v} + \sum_{i=1}^T {c}_i \log {y}_i^{a} ),
\end{equation}

\begin{equation}
\mathcal{L}_{\text {focal}}^{mul}=-\sum_{i=1}^T {c}_i \log {y}_i^{m},
\end{equation}
where ${y}_i^{a}$, ${y}_i^{v}$, and ${y}_i^{m}$ represent the predicted outcomes for audio, visual, and multi-modal fusion, respectively.

To further promote inter-modality fusion and ensure stability in the joint optimization process, we introduce an alignment loss to align the predictions from multiple branches towards consistency. This loss adopts the Kullback-Leibler divergence loss (KLDivLoss)~\cite{wang2023cocaclip} to align the probability distributions of uni-modal predictions with the multi-modal fusion predictions.

\begin{equation}
\mathcal{L}_{\text{KL}}^{v \rightarrow m} = \sum_{i=1}^T {y}_i^{v} \log \left( \frac{{y}_i^{v}}{{y}_i^{m}} \right),
\end{equation}

\begin{equation}
\mathcal{L}_{\text{KL}}^{a \rightarrow m} = \sum_{i=1}^T {y}_i^{a} \log \left( \frac{{y}_i^{a}}{{y}_i^{m}} \right).
\end{equation}

\paragraph{\textbf{Training Objective.}} 
To more accurately identify the clips that should be selected, we employ a customized 1-D IoU-based loss~\cite{rezatofighi2019generalized}, denoted as \(\mathcal{L}_{iou}\), which refines our temporal predictions. This approach is similar to a variant of the Temporal Event Localization framework~\cite{zhang2022actionformer} and is related to the task of audio-visual event detection~\cite{geng2023dense}. To achieve this goal, we design a regression head with a structure similar to that of the classification head, but incorporating a ReLU activation layer at its final stage.

Formally, the composite objective function for a given video is represented as follows:
\begin{equation}
\lambda_1\mathcal{L}_{\text {focal}}^{uni}+\lambda_2\mathcal{L}_{\text {focal}}^{mul}+\lambda_3(\mathcal{L}_{\text{KL}}^{v \rightarrow m}+\mathcal{L}_{\text{KL}}^{a \rightarrow m})+\lambda_4\mathcal{L}_{iou},
\end{equation}
where the hyper-parameters \(\lambda_{1-4}\) are equilibrium constants.

\section{Experiments}
\subsection{Setting}
\paragraph{\textbf{Feature Extraction.}} 
Each video is sampled at one frame per second. Each frame is then processed using the CLIP ViT-B/32 model~\cite{radford2021learning} to derive 512-dimensional visual features. In parallel, each one-second audio segment is input into a pre-trained PANN model~\cite{kong2020panns} trained on AudioSet~\cite{gemmeke2017audio}, yielding 2048-dimensional audio features. For caption features, we use the WhisperX~\cite{bain2022whisperx} model for Automatic Speech Recognition (ASR) with timestamps, followed by the all-MiniLM-L6-v2 model from Sentence-Transformers~\cite{reimers2019sentence} to extract 384-dimensional sentence features. We associate transcript sentences with corresponding video segments to address the temporal misalignment between transcripts and video frames. If a sentence overlaps multiple segments, we duplicate it to create segment-transcript pairs. For empty captions within a time window, we use a special [empty] token as a placeholder. Through these steps, visual, audio, and caption features align at the segment level.

\paragraph{\textbf{Implementation Details.}} 
We partition the dataset into train/val/test splits at a ratio of 8/1/1. The embedding dimension of the model is set to \(d = 512\), and the number of layers \(N_s\), \(N_c\), and \(N_f\) is set to 3. We utilize the Adam optimizer with a learning rate of 1e-4 for 100 epochs, which is adjusted using cosine learning rate decay, while the first 5 epochs employ linear warm-up to facilitate stable learning. The hyper-parameters \(\lambda_{1-4}\) are set to 0.1, 0.3, 0.1, and 0.7, respectively. All experiments are conducted on two A100 GPUs within the PyTorch framework.

\begin{table}[t]
    \centering
    \setlength\tabcolsep{1mm} 
    \fontsize{9}{11}\selectfont
    \begin{tabular}{cccccccc}
        \toprule
        Method & Modality & 0.5 & 0.6 & 0.7 & 0.8 & 0.9 & Avg.  \\
        \midrule
         SL-Module & V & 21.89 & 14.11 & \underline{8.48} & \underline{3.91} & 0.97 & 9.87 \\
         Moment-DETR & V & 23.29 & \underline{14.70} & 8.25 & 3.45 & 0.98 & 10.13 \\
         QD-DETR & V & 23.02 & 14.47 & 8.01 & 3.54 & 0.99 &  10.01 \\
         UMT & V\&A & 21.46 & 13.06 & 7.52 & 3.49 & \underline{1.04} & 9.31 \\
         QD-DETR & V\&A & \underline{23.30} & 14.64 & 7.98 & 3.87 & 1.01 & \underline{10.16} \\
         Ours & V\&A\&C & \textbf{24.94} &  \textbf{16.35}& \textbf{10.14} &\textbf{4.86}& \textbf{1.57} & \textbf{11.57} \\  
        \bottomrule
    \end{tabular}
        \caption{Comparison with State-of-the-Art Temporal Grounding Models: SL-Module~\cite{xu2021cross}, Moment-DETR~\cite{lei2021detecting}, UMT~\cite{liu2022umt}, and QD-DETR~\cite{moon2023query}. The numbers in the first row of the table represent the thresholds. To denote the best results, we use bold text, and for the second-best results, we underline them.}
    \label{tab:main_results}
\end{table}

\paragraph{\textbf{Evaluation Metrics.}} We initially apply a threshold of 0.5 to filter out segments with low confidence. Subsequently, we employ soft-nms~\cite{bodla2017soft} to obtain the top-k clips. The value of \(k\) is dynamically determined based on statistical results from the training set, where we set it as 3 clips every ten minutes on average. We present the results using tIoU with thresholds of [0.5, 0.9, 0.1].

\subsection{Results and Analysis}
\paragraph{\textbf{Comparison With Baselines.}} We approach the task of video repurposing as a temporal grounding problem with multi-modal inputs, though without an explicit query. For comparison, we select several representative temporal grounding models as baselines~\cite{xu2021cross,lei2021detecting,liu2022umt,moon2023query}. The results, presented in Table \ref{tab:main_results}, show that our model significantly outperforms these baselines across various IoU thresholds, consistently achieving top scores and demonstrating superior performance.

\begin{table}[t]
    \centering
    \setlength\tabcolsep{1mm} 
    \fontsize{9}{11}\selectfont
    \begin{tabular}{ccccccccc}
        \toprule
        Modality & UF & AL & 0.5 & 0.6 & 0.7 & 0.8 & 0.9 & Avg.  \\
        \midrule
         A & - & - & 22.20 & 13.99 & 8.13 & 3.34 & 0.63&  9.66 \\
         V & - & - & 21.91 & 13.87 & 8.08 & 3.14 & 0.60&  9.52 \\
         C & - & - & 22.30 & 14.02 & 7.98 & 3.06 & 0.57 &  9.59 \\
         A\&V & - & - & 23.28 & 14.95 & 8.69 & 3.71 & 0.90 &  10.31\\
            A\&V & \ding{51} &  - & 23.90 & 15.54 & 9.10 & 3.97 & 1.02 & 10.71 \\
            A\&V & \ding{51} & \ding{51} & 24.13 & 15.71 & 9.47 & 4.30 & 1.21 & 10.96 \\
          A\&V\&C & - & - & 23.74 &  15.47& 9.13 &4.26& 1.19 & 10.75 \\
            A\&V\&C & \ding{51} & - & 24.51 & 15.97 & 9.86 & 4.63 & 1.45 & 11.28 \\
          A\&V\&C & \ding{51} & \ding{51} & \textbf{24.94} &  \textbf{16.35}& \textbf{10.14} &\textbf{4.86}& \textbf{1.57} & \textbf{11.57} \\
        \bottomrule
    \end{tabular}
        \caption{Ablation study of the impact of various modalities and the primary results on the Repurposing-10K test set. A: Audio modality. V: Visual modality. A\&V: Audio modality combined with visual modality. A\&V\&C: Audio modality and visual modality integrated with caption modality. UF: Uni-modal focal loss. AL: Alignment loss. The numbers in the first row of the table represent the thresholds.}
    \label{tab:ablation}
\end{table}

\begin{figure*}[t]
    \centering
    \includegraphics[width=0.9\linewidth]{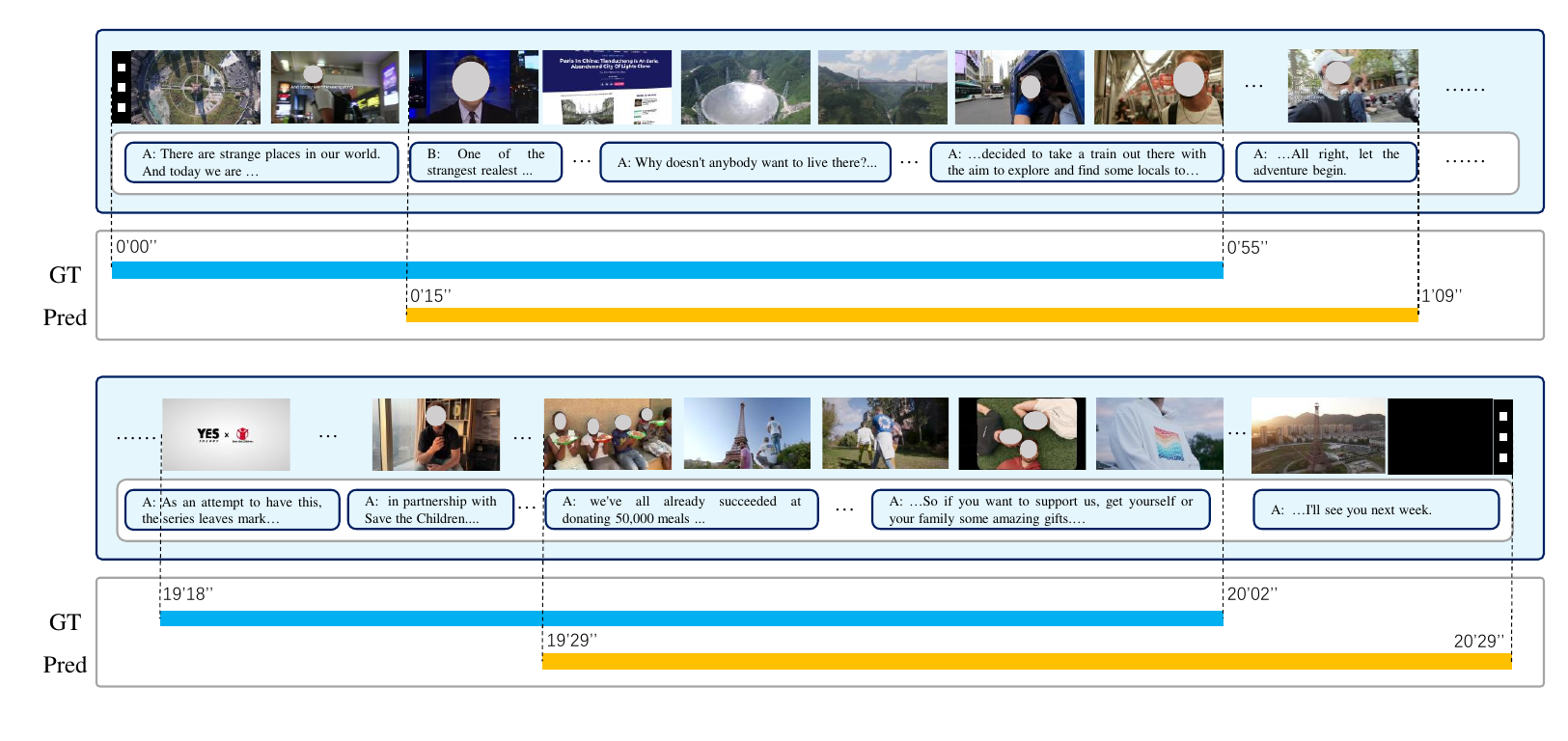}
    \caption{Two visual examples of video repurposing results. Blue: Ground Truth. Yellow: Predictions.}
    \label{fig:results}
\end{figure*}

\paragraph{\textbf{Modality Fusion Ablations.}} 
We first examine the impact of different modalities on the model's performance. As shown in Table~\ref{tab:ablation}, in uni-modal scenarios, the model using caption features achieves the best average performance. Notably, it outperforms the visual-based model at lower thresholds, highlighting the effectiveness of caption features when a single modality is considered. Additionally, performance consistently improves when both audio and visual features are utilized together. This improvement is further enhanced with the integration of the Caption Enhancement Module, underscoring the importance of multimodal learning for the video repurposing task.

\paragraph{\textbf{Modality Alignment Ablations.}} 
We also investigate the impact of the Multi-Modal Align Guider module. As shown in Table~\ref{tab:ablation}, incorporating auxiliary multimodal constraints enhances the model's robustness. Allowing uni-modal branches to make independent predictions can improve performance to some extent. However, this improvement is significantly amplified when their distributions are explicitly aligned with multimodal predictions. This suggests that the alignment loss not only indirectly reinforces information from individual modalities but also directly guides the multimodal branches to learn more effectively.

\paragraph{\textbf{Modality Fusion Layers Ablations.}} 
To assess the impact of different feature fusion depths on model performance, we experiment with varying the number of layers in distinct modules, denoted as \(N_s\), \(N_c\), and \(N_f\). To ensure a fair comparison, the total number of layers across these modules is kept constant at nine. As shown in Table \ref{tab:fusion}, a balanced configuration of 3/3/3 layers produces the best results. Interestingly, highly imbalanced configurations, such as 7/1/1, perform worse than models using a single modality. This highlights the importance of not only considering interactions within individual modalities but also fostering effective cross-modal interactions. Such an approach is essential for fully leveraging the potential of features across different modalities. Additionally, our findings indicate that shallow layers for attention mechanisms are insufficient for the model to establish meaningful connections between various modalities.

\begin{table}[t]
    \centering
    \setlength\tabcolsep{1mm} 
    \fontsize{9}{11}\selectfont
    \begin{tabular}{ccccccc}
        \toprule
        \(N_s\)/\(N_c\)/\(N_f\) & 0.5 & 0.6 & 0.7 & 0.8 & 0.9 & Avg.  \\
        \midrule
        1 / 4 / 4 & 23.37 & 15.01 & 8.62 & 4.10 & 1.29 & 10.48 \\
         3 / 3 / 3 & \textbf{24.94} &  \textbf{16.35}& \textbf{10.14} &\textbf{4.86}& \textbf{1.57} & \textbf{11.57} \\
        5 / 2 / 2 & 22.51 & 13.60 & 7.23 & 3.25 & 0.80 &  9.48 \\
        7 / 1 / 1 & 19.88  & 12.41 & 6.77 & 3.08& 0.73 &  8.57 \\
        \bottomrule
    \end{tabular}
        \caption{Ablation study to explore the selection of various numbers of transformer modality fusion layers. \(N_s\): Number of self-attention layers. \(N_c\): Cross-attention layer for caption enhancement. \(N_f\): Fusion layer for integrating audio and visual features. The numbers in the first row of the table represent the thresholds.}
    \label{tab:fusion}
\end{table}

\paragraph{\textbf{Qualitative Results.}} 
To further demonstrate the effectiveness of our model, we present two practical applications as illustrated in Figure~\ref{fig:results}. In both instances, there is a notable alignment between our model's predictions and the actual ground truth. In the first scenario, the ground truth annotation provides a straightforward and concise description, leading to the decision to explore these “strange places.” In contrast, our model's prediction starts with an engaging B-roll news clip and ends with the main character's declaration, “Let the adventure begin.” In the second scenario, while the annotation offers a detailed introduction to the product of the video creator, our predictions more effectively highlight the video's concluding section. Both selected clips begin with intriguing content and end with fitting remarks, demonstrating their ability to engage and provide a complete narrative.

\section{Conclusion}
In this paper, we curate the Repurpose-10K dataset for the challenging task of video repurposing. The dataset comprises over 10,000 videos and more than 120,000 clips, annotated with User Generated Content (UGC) in the real world. Furthermore, we present an end-to-end model designed to address the video repurposing challenge by jointly learning classification and regression tasks. Through extensive experiments and ablation studies, we demonstrate the effectiveness of the proposed model.

\section{Acknowledgments}
This work is supported by the National Science Foundation of China (62206048), the Natural Science
Foundation of Jiangsu Province (BK20220819), and the Fundamental Research Funds for the Central Universities (2242024k30035). The Big Data Computing Center of Southeast University also supports this research work.

\bibliography{aaai25}

\end{document}